# PixelCNN++: Improving the PixelCNN with Discretized Logistic Mixture Likelihood and Other Modifications

**Tim Salimans, Andrej Karpathy, Xi Chen, Diederik P. Kingma**
{tim,karpathy,peter,dpkingma}@openai.com

## Abstract

PixelCNNs are a recently proposed class of powerful generative models with tractable likelihood. Here we discuss our implementation of PixelCNNs which we make available at https://github.com/openai/pixel-cnn. Our implementation contains a number of modifications to the original model that both simplify its structure and improve its performance. 1) We use a discretized logistic mixture likelihood on the pixels, rather than a 256-way softmax, which we find to speed up training. 2) We condition on whole pixels, rather than R/G/B sub-pixels, simplifying the model structure. 3) We use downsampling to efficiently capture structure at multiple resolutions. 4) We introduce additional short-cut connections to further speed up optimization. 5) We regularize the model using dropout. Finally, we present state-of-the-art log likelihood results on CIFAR-10 to demonstrate the usefulness of these modifications.

## 1 Introduction

The PixelCNN, introduced by van den Oord et al. (2016b), is a generative model of images with a tractable likelihood. The model fully factorizes the probability density function on an image **x** over all its sub-pixels (color channels in a pixel) as $p(\mathbf{x}) = \prod_i p(x_i|x_{<i})$. The conditional distributions $p(x_i|x_{<i})$ are parameterized by convolutional neural networks and all share parameters. The PixelCNN is a powerful model as the functional form of these conditionals is very flexible. In addition it is computationally efficient as all conditionals can be evaluated in parallel on a GPU for an observed image **x**. Thanks to these properties, the PixelCNN represents the current state-of-the-art in generative modeling when evaluated in terms of log-likelihood. Besides being used for modeling images, the PixelCNN model was recently extended to model audio (van den Oord et al., 2016a), video (Kalchbrenner et al., 2016b) and text (Kalchbrenner et al., 2016a).

For use in our research, we developed our own internal implementation of PixelCNN and made a number of modifications to the base model to simplify its structure and improve its performance. We now release our implementation at https://github.com/openai/pixel-cnn, hoping that it will be useful to the broader community. Our modifications are discussed in Section 2, and evaluated experimentally in Section 3. State-of-the-art log-likelihood results confirm their usefulness.

## 2 Modifications to PixelCNN

We now describe the most important modifications we have made to the PixelCNN model architecure as described by van den Oord et al. (2016c). For complete details see our code release at https://github.com/openai/pixel-cnn.

### 2.1 Discretized logistic mixture likelihood

The standard PixelCNN model specifies the conditional distribution of a sub-pixel, or color channel of a pixel, as a full 256-way softmax. This gives the model a lot of flexibility, but it is also very costly in terms of memory. Moreover, it can make the gradients with respect to the network parameters very sparse, especially early in training. With the standard parameterization, the model does not know that a value of 128 is close to a value of 127 or 129, and this relationship first has to be learned before the model can move on to higher level structures. In the extreme case where a particular sub-pixel value is never observed, the model will learn to assign it zero probability. This would be especially problematic for data with higher accuracy on the observed pixels than the usual 8 bits: In the extreme case where very high precision values are observed, the PixelCNN, in its current form, would require a prohibitive amount of memory and computation, while learning very slowly. We therefore propose a different mechanism for computing the conditional probability of the observed discretized pixel values. In our model, like in the VAE of Kingma et al. (2016), we assume there is a latent color intensity $\nu$ with a continuous distribution, which is then rounded to its nearest 8-bit representation to give the observed sub-pixel value $x$. By choosing a simple continuous distribution for modeling $\nu$ (like the logistic distribution as done by Kingma et al. (2016)) we obtain a smooth and memory efficient predictive distribution for $x$. Here, we take this continuous univariate distribution to be a mixture of logistic distributions which allows us to easily calculate the probability on the observed discretized value $x$, as shown in equation (2). For all sub-pixel values $x$ excepting the edge cases 0 and 255 we have:

$$\nu \sim \sum_{i=1}^{K} \pi_i \text{logistic}(\mu_i, s_i) \tag{1}$$

$$P(x|\pi, \mu, s) = \sum_{i=1}^{K} \pi_i \left[\sigma((x + 0.5 - \mu_i)/s_i) - \sigma((x - 0.5 - \mu_i)/s_i)\right], \tag{2}$$

where $\sigma()$ is the logistic sigmoid function. For the edge case of 0, replace $x - 0.5$ by $-\infty$, and for 255 replace $x + 0.5$ by $+\infty$. Our provided code contains a numerically stable implementation for calculating the log of the probability in equation 2.

Our approach follows earlier work using continuous mixture models (Domke et al., 2008; Theis et al., 2012; Uria et al., 2013; Theis & Bethge, 2015), but avoids allocating probability mass to values outside the valid range of $[0, 255]$ by explicitly modeling the rounding of $\nu$ to $x$. In addition, we naturally assign higher probability to the edge values 0 and 255 than to their neighboring values, which corresponds well with the observed data distribution as shown in Figure 1. Experimentally, we find that only a relatively small number of mixture components, say 5, is needed to accurately model the conditional distributions of the pixels. The output of our network is thus of much lower dimension, yielding much denser gradients of the loss with respect to our parameters. In our experiments this greatly sped up convergence during optimization, especially early on in training. However, due to the other changes in our architecture compared to that of van den Oord et al. (2016c) we cannot say with certainty that this would also apply to the original PixelCNN model.

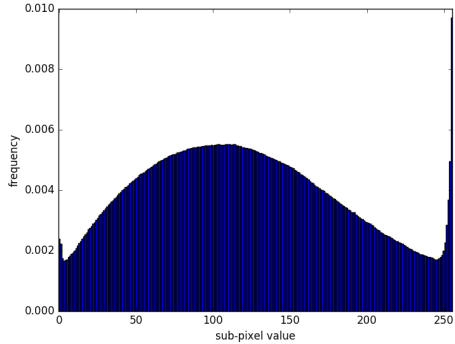

Figure 1: Marginal distribution of all sub-pixel values in CIFAR-10. The edge value of 255 is much more frequent than its neighbouring values: This is easy to model using our rounding based approach, but harder using continuous or truncated distributions.

## 2.2 Conditioning on whole pixels

The pixels in a color image consist of three real numbers, giving the intensities of the red, blue and green colors. The original PixelCNN factorizes the generative model over these 3 sub-pixels. This allows for very general dependency structure, but it also complicates the model: besides keeping track of the spatial location of feature maps, we now have to separate out all feature maps in 3 groups depending on whether or not they can see the R/G/B sub-pixel of the current location. This added complexity seems to be unnecessary as the dependencies between the color channels of a pixel are likely to be relatively simple and do not require a deep network to model. Therefore, we instead condition only on whole pixels up and to the left in an image, and output joint predictive distributions over all 3 channels of a predicted pixel. The predictive distribution on a pixel itself can be interpreted as a simple factorized model: We first predict the red channel using a discretized mixture of logistics as described in section 2.1. Next, we predict the green channel using a predictive distribution of the same form. Here we allow the means of the mixture components to linearly depend on the value of the red sub-pixel. Finally, we model the blue channel in the same way, where we again only allow linear dependency on the red and green channels. For the pixel $(r_{i,j}, g_{i,j}, b_{i,j})$ at location $(i,j)$ in our image, the distribution conditional on the context $C_{i,j}$, consisting of the mixture indicator and the previous pixels, is thus

$$\begin{aligned} p(r_{i,j}, g_{i,j}, b_{i,j} | C_{i,j}) &= P(r_{i,j} | \mu_r(C_{i,j}), s_r(C_{i,j})) \times P(g_{i,j} | \mu_g(C_{i,j}, r_{i,j}), s_g(C_{i,j})) \\ &\quad \times P(b_{i,j} | \mu_b(C_{i,j}, r_{i,j}, g_{i,j}), s_b(C_{i,j})) \\ \mu_g(C_{i,j}, r_{i,j}) &= \mu_g(C_{i,j}) + \alpha(C_{i,j}) r_{i,j} \\ \mu_b(C_{i,j}, r_{i,j}, g_{i,j}) &= \mu_b(C_{i,j}) + \beta(C_{i,j}) r_{i,j} + \gamma(C_{i,j}) b_{i,j}, \end{aligned} \quad (3)$$

with $\alpha, \beta, \gamma$ scalar coefficients depending on the mixture component and previous pixels.

The mixture indicator is shared across all 3 channels; i.e. our generative model first samples a mixture indicator for a pixel, and then samples the color channels one-by-one from the corresponding mixture component. Had we used a discretized mixture of univariate Gaussians for the sub-pixels, instead of logistics, this would have been exactly equivalent to predicting the complete pixel using a (discretized) mixture of 3-dimensional Gaussians with full covariance. The logistic and Gaussian distributions are very similar, so this is indeed very close to what we end up doing. For full implementation details we refer to our code at https://github.com/openai/pixel-cnn.

## 2.3 Downsampling versus dilated convolution

The original PixelCNN only uses convolutions with small receptive field. Such convolutions are good at capturing local dependencies, but not necessarily at modeling long range structure. Although we find that capturing these short range dependencies is often enough for obtaining very good log-likelihood scores (see Table 2), explicitly encouraging the model to capture long range dependencies can improve the perceptual quality of generated images (compare Figure 3 and Figure 5). One way of allowing the network to model structure at multiple resolutions is to introduce dilated convolutions into the model, as proposed by van den Oord et al. (2016a) and Kalchbrenner et al. (2016b). Here, we instead propose to use downsampling by using convolutions of stride 2. Downsampling accomplishes the same multi-resolution processing afforded by dilated convolutions, but at a reduced computational cost: where dilated convolutions operate on input of ever increasing size (due to zero padding), downsampling reduces the input size by a factor of 4 (for stride of 2 in 2 dimensions) at every downsampling. The downside of using downsampling is that it loses information, but we can compensate for this by introducing additional short-cut connections into the network as explained in the next section. With these additional short-cut connections, we found the performance of downsampling to be the same as for dilated convolution.

## 2.4 Adding short-cut connections

For input of size $32 \times 32$ our suggested model consists of 6 blocks of 5 ResNet layers. In between the first and second block, as well as the second and third block, we perform subsampling by strided convolution. In between the fourth and fifth block, as well as the fifth and sixth block, we perform upsampling by transposed strided convolution. This subsampling and upsampling process loses information, and we therefore introduce additional short-cut connections into the model to recover this information from lower layers in the model. The short-cut connections run from the ResNet layers in the first block to the corresponding layers in the sixth block, and similarly between blocks two and five, and blocks three and four. This structure resembles the VAE model with top down inference used by Kingma et al. (2016), as well as the U-net used by Ronneberger et al. (2015) for image segmentation. Figure 2 shows our model structure graphically.

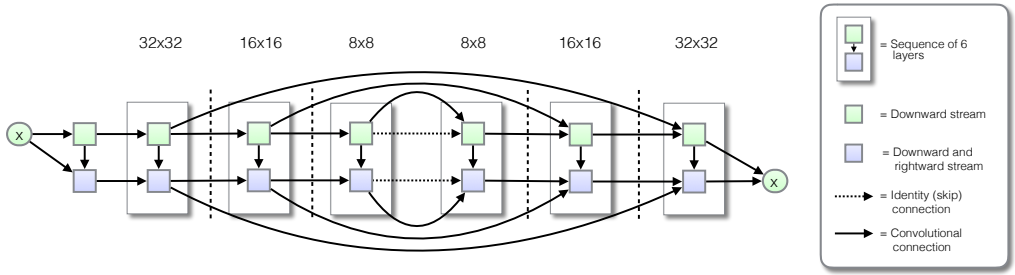

Figure 2: Like van den Oord et al. (2016c), our model follows a two-stream (downward, and downward+rightward) convolutional architecture with residual connections; however, there are two significant differences in connectivity. First, our architecture incorporates downsampling and upsampling, such that the inner parts of the network operate over larger spatial scale, increasing computational efficiency. Second, we employ long-range skip-connections, such that each $k$-th layer provides a direct input to the $(K - k)$-th layer, where $K$ is the total number of layers in the network. The network is grouped into sequences of six layers, where most sequences are separated by downsampling or upsampling.

### 2.5 REGULARIZATION USING DROPOUT

The PixelCNN model is powerful enough to overfit on training data. Moreover, rather than just reproducing the training images, we find that overfitted models generate images of low perceptual quality, as shown in Figure 8. One effective way of regularizing neural networks is dropout (Srivastava et al., 2014). For our model, we apply standard binary dropout on the residual path after the first convolution. This is similar to how dropout is applied in the *wide residual networks* of Zagoruyko & Komodakis (2016). Using dropout allows us to successfully train high capacity models while avoiding overfitting and producing high quality generations (compare figure 8 and figure 3).

## 3 EXPERIMENTS

We apply our model to modeling natural images in the CIFAR-10 data set. We achieve state-of-the-art results in terms of log-likelihood, and generate images with coherent global structure.

### 3.1 UNCONDITIONAL GENERATION ON CIFAR-10

We apply our PixelCNN model, with the modifications as described above, to generative modeling of the images in the CIFAR-10 data set. For the encoding part of the PixelCNN, the model uses 3 Resnet blocks consisting of 5 residual layers, with $2 \times 2$ downsampling in between. The same architecture is used for the decoding part of the model, but with upsampling instead of downsampling in between blocks. All residual layers use 192 feature maps and a dropout rate of $0.5$. Table 1 shows the state-of-the-art test log-likelihood obtained by our model. Figure 3 shows some samples generated by the model.

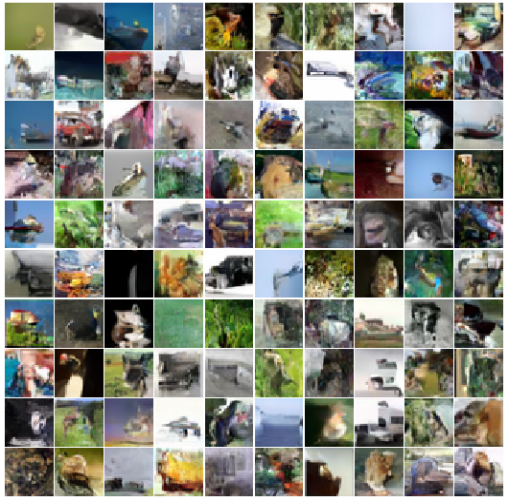

Figure 3: Samples from our PixelCNN model trained on CIFAR-10.

| Model | Bits per sub-pixel |
|---|---|
| Deep Diffusion (Sohl-Dickstein et al., 2015) | 5.40 |
| NICE (Dinh et al., 2014) | 4.48 |
| DRAW (Gregor et al., 2015) | 4.13 |
| Deep GMMs (van den Oord & Dambre, 2015) | 4.00 |
| Conv DRAW (Gregor et al., 2016) | 3.58 |
| Real NVP (Dinh et al., 2016) | 3.49 |
| PixelCNN (van den Oord et al., 2016b) | 3.14 |
| VAE with IAF (Kingma et al., 2016) | 3.11 |
| Gated PixelCNN (van den Oord et al., 2016c) | 3.03 |
| PixelRNN (van den Oord et al., 2016b) | 3.00 |
| **PixelCNN++** | **2.92** |

Table 1: Negative log-likelihood for generative models on CIFAR-10 expressed as bits per sub-pixel.

## 3.2 CLASS-CONDITIONAL GENERATION

Next, we follow van den Oord et al. (2016c) in making our generative model conditional on the class-label of the CIFAR-10 images. This is done by linearly projecting a one-hot encoding of the class-label into a separate class-dependent bias vector for each convolutional unit in our network. We find that making the model class-conditional makes it harder to avoid overfitting on the training data: our best test log-likelihood is **2.94** in this case. Figure 4 shows samples from the class-conditional model, with columns 1-10 corresponding the 10 classes in CIFAR-10. The images clearly look qualitatively different across the columns and for a number of them we can clearly identify their class label.

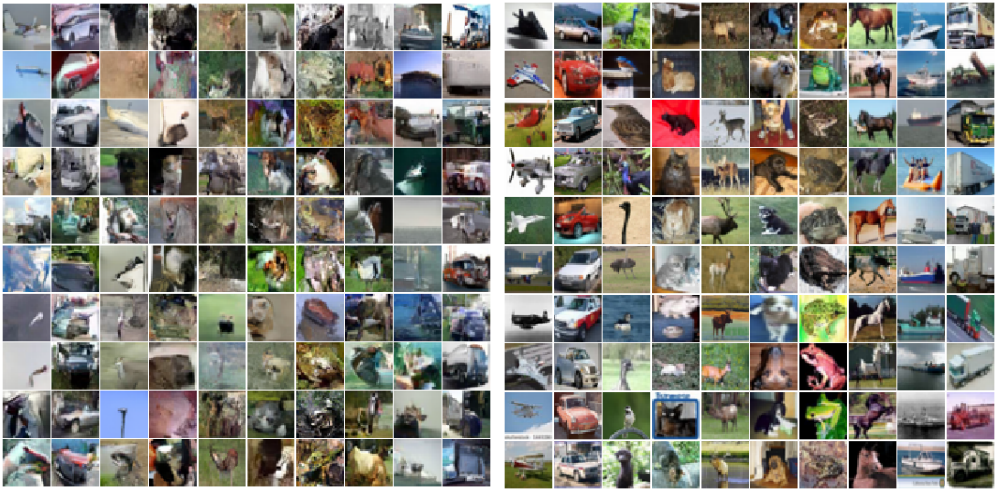

Figure 4: Class-conditional samples from our PixelCNN for CIFAR-10 (left) and real CIFAR-10 images for comparison (right).

### 3.3 Examining network depth and field of view size

It is hypothesized that the size of the receptive field and additionally the removal of blind spots in the receptive field are important for PixelCNN's performance (van den Oord et al., 2016b). Indeed van den Oord et al. (2016c) specifically introduced an improvement over the previous PixelCNN model to remove the blind spot in the receptive field that was present in their earlier model.

Here we present the surprising finding that in fact a PixelCNN with rather small receptive field can attain competitive generative modelling performance on CIFAR-10 as long as it has enough capacity. Specifically, we experimented with our proposed PixelCNN++ model without downsampling blocks and reduce the number of layers to limit the receptive field size. We investigate two receptive field sizes: 11x5 and 15x8, and a receptive field size of 11x5, for example, means that the conditional distribution of a pixel can depends on a rectangle above the pixel of size 11x5 as well as $\frac{11-1}{2} = 5$x1 block to the left of the pixel.

As we limit the size of the receptive field, the capacity of the network also drops significantly since it contains many fewer layers than a normal PixelCNN. We call the type of PixelCNN that's simply limited in depth "Plain" Small PixelCNN. Interestingly, this model already has better performance than the original PixelCNN in van den Oord et al. (2016b) which had a blind spot. To increase capacity, we introduced two simple variants that make Small PixelCNN more expressive without growing the receptive field:

- NIN (Network in Network): insert additional gated ResNet blocks with 1x1 convolution between regular convolution blocks that grow receptive field. In this experiment, we inserted 3 NIN blocks between every other layer.
- Autoregressive Channel: skip connections between sets of channels via 1x1 convolution gated ResNet block.

Both modifications increase the capacity of the network, resulting in improved log-likelihood as shown in Table 2. Although the model with small receptive field already achieves an impressive likelihood score, its samples do lack global structure, as seen in Figure 5.

Table 2: CIFAR-10 bits per sub-pixel for Small PixelCNN

| Model | Bits per sub-pixel |
|---|---|
| Field=11x5, Plain | 3.11 |
| Field=11x5, NIN | 3.09 |
| Field=11x5, Autoregressive Channel | 3.07 |
| Field=15x8, Plain | 3.07 |
| Field=15x8, NIN | 3.04 |
| Field=15x8, Autoregressive Channel | 3.03 |

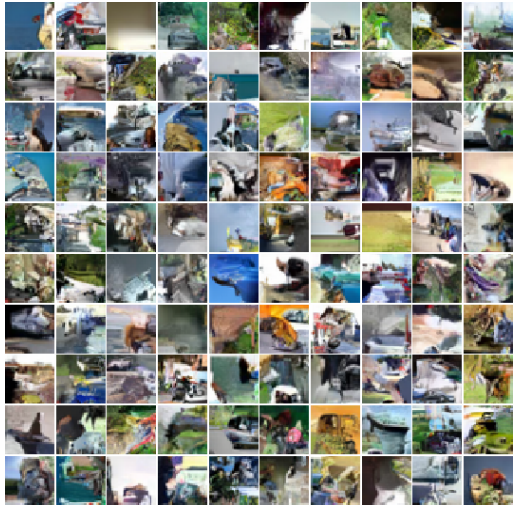

Figure 5: Samples from 3.03 bits/dim Small PixelCNN

### 3.4 Ablation experiments

In order to test the effect of our modifications to PixelCNN, we run a number of ablation experiments where for each experiment we remove a specific modification.

#### 3.4.1 Softmax likelihood instead of discretized logistic mixture

In order to test the contribution of our logistic mixture likelihood, we re-run our CIFAR-10 experiment with the 256-way softmax as the output distribution instead. We allow the 256 logits for each sub-pixel to linearly depend on the observed value of previous sub-pixels, with coefficients that are given as output by the model. Our model with softmax likelihood is thus strictly more flexible than our model with logistic mixture likelihood, although the parameterization is quite different from that used by van den Oord et al. (2016c). The model now outputs 1536 numbers per pixel, describing the logits on the 256 potential values for each sub-pixel, as well as the coefficients for the dependencies between the sub-pixels. Figure 6 shows that this model trains more slowly than our original model. In addition, the running time per epoch is significantly longer for our tensorflow implementation. For our architecture, the logistic mixture model thus clearly performs better. Since our architecture differs from that of van den Oord et al. (2016c) in other ways as well, we cannot say whether this would also apply to their model.

#### 3.4.2 Continuous mixture likelihood instead of discretization

Instead of directly modeling the discrete pixel values in an image, it is also possible to *de-quantize* them by adding noise from the standard uniform distribution, as used by Uria et al. (2013) and others, and modeling the data as being continuous. The resulting model can be interpreted as a variational autoencoder (Kingma & Welling, 2013; Rezende et al., 2014), where the dequantized pixels $\mathbf{z}$ form a latent code whose prior distribution is captured by our model. Since the original discrete pixels $\mathbf{x}$ can be perfectly reconstructed from $\mathbf{z}$ under this model, the usual reconstruction term vanishes from the variational lower bound. The entropy of the standard uniform distribution is zero, so the term that remains is the log likelihood of the dequantized pixels, which thus gives us a variational lower bound on the log likelihood of our original data.

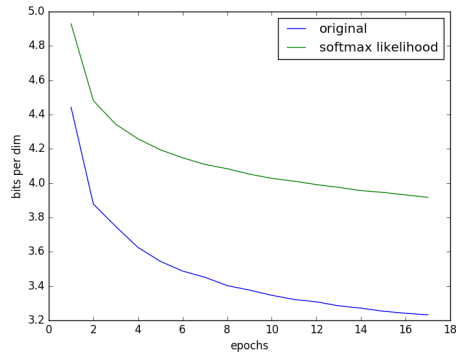

Figure 6: Training curves for our model with logistic mixture likelihood versus our model with softmax likelihood.

We re-run our model for CIFAR-10 using the same model settings as those used for the 2.92 bits per dimension result in Table 1, but now we remove the discretization in our likelihood model and instead add standard uniform noise to the image data. The resulting model is a continuous mixture model in the same class as that used by Theis et al. (2012); Uria et al. (2013); Theis & Bethge (2015) and others. After optimization, this model gives a variational lower bound on the data log likelihood of 3.11 bits per dimension. The difference with the reported 2.92 bits per dimension shows the benefit of using discretization in the likelihood model.

### 3.4.3 NO SHORT-CUT CONNECTIONS

Next, we test the importance of the additional parallel short-cut connections in our model, indicated by the dotted lines in Figure 2. We re-run our unconditional CIFAR-10 experiment, but remove the short-cut connections from the model. As seen in Figure 7, the model fails to train without these connections. The reason for needing these extra short-cuts is likely to be our use of sub-sampling, which discards information that otherwise cannot easily be recovered,

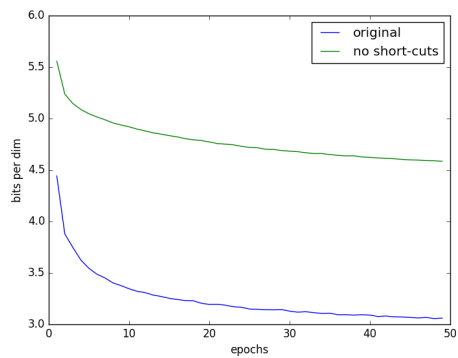

Figure 7: Training curves for our model with and without short-cut connections.

### 3.4.4 NO DROPOUT

We re-run our CIFAR-10 model without dropout regularization. The log-likelihood we achieve on the training set is below 2.0 bits per sub-pixel, but the final test log-likelihood is above 6.0 bits per sub-pixel. At no point during training does the unregularized model get a test-set log-likelihood below 3.0 bits per sub-pixel. Contrary to what we might naively expect, the perceptual quality of the generated images by the overfitted model is not great, as shown in Figure 8.

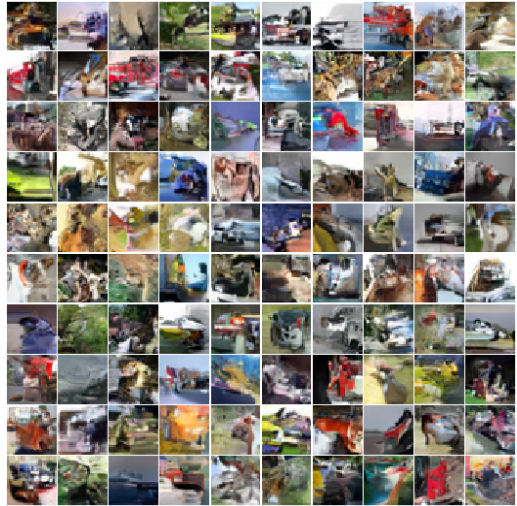

Figure 8: Samples from intentionally overfitted PixelCNN model trained on CIFAR-10, with train log-likelihood of 2.0 bits per dimension: Overfitting does not result in great perceptual quality.

## 4 CONCLUSION

We presented PixelCNN++, a modification of PixelCNN using a discretized logistic mixture likelihood on the pixels among other modifications. We demonstrated the usefulness of these modifications with state-of-the-art results on CIFAR-10. Our code is made available at https://github.com/openai/pixel-cnn and can easily be adapted for use on other data sets.